\documentclass[pdflatex,sn-basic]{sn-jnl}%

\usepackage{xfrac}
\usepackage{ulem}
\usepackage{pifont}
\usepackage{multirow}
\usepackage{makecell}

\jyear{2022}%

\begin{document}

\title[A DNN Framework for Learning Lagrangian Drift With Uncertainty]{A DNN Framework for Learning Lagrangian Drift With Uncertainty}

\author*[1,2,3]{\fnm{Joseph} \sur{Jenkins}}\email{joseph.jenkins@lis-lab.fr}
\author[2]{\fnm{Adeline} \sur{Paiement}}
\author[3]{\fnm{Yann} \sur{Ourmières}}
\author[4]{\fnm{Julien} \sur{Le Sommer}}
\author[5]{\fnm{Jacques} \sur{Verron}}
\author[1]{\fnm{Clément} \sur{Ubelmann}}
\author[2]{\fnm{Hervé} \sur{Glotin}}

\affil[1]{Datlas, France}
\affil[2]{Université de Toulon, Aix Marseille Univ, CNRS, LIS, Marseille, France}
\affil[3]{Mediterranean Institute of Oceanography, Université de Toulon, Aix Marseille Univ, CNRS, IRD, MIO, Toulon, France}
\affil[4]{Univ. Grenoble Alpes, CNRS, IRD, Grenoble INP, IGE, 38000 Grenoble, France}
\affil[5]{Hydro Matters, France}

\abstract{Reconstructions of Lagrangian drift, for example for objects lost at sea, are often uncertain due to unresolved physical phenomena within the data. Uncertainty is usually overcome by introducing stochasticity into the drift, but this approach requires specific assumptions for modelling uncertainty. We remove this constraint by presenting a purely data-driven framework for modelling probabilistic drift in flexible environments. Using ocean circulation model simulations, we generate probabilistic trajectories of object location by simulating uncertainty in the initial object position. We train an emulator of probabilistic drift over one day given perfectly known velocities and observe good agreement with numerical simulations. Several loss functions are tested. Then, we strain our framework by training models where the input information is imperfect. On these harder scenarios, we observe reasonable predictions although the effects of data drift become noticeable when evaluating the models against unseen flow scenarios. Source code and data is available at \url{https://github.com/JenkinsJR/UncertainDrift}.}

\maketitle

\section{Introduction}
\label{sec:intro}

We present a data-driven framework for learning Lagrangian drift over a given timestep (e.g.~one day in our experiments) in the presence of uncertainty. Uncertainty arises when dynamical systems cannot be perfectly described due to imperfect modelling of either the dynamics or the system state. Modelling capabilities are constrained by availability of compute, knowledge of the underlying physical phenomena (particularly at small scales), and sensor resolution. Meanwhile, chaotic systems result in minor state variations to have significant and unpredictable influences in the observed behaviour. Thus, modelling drift with uncertainty is critical for many applications whose dynamics are chaotic or whose inputs (e.g. velocity field) do not sufficiently resolve the necessary dynamics. This includes applications such as ocean and atmospheric dynamics~\cite{gill1982}.

Uncertainty is usually modelled through stochastic differential equations (SDEs) to account for uncaptured physical phenomena at small scales~\cite{van2018}. If the phenomena can be captured such that the main source of uncertainty comes from the effects of chaos, accounting for uncertainty may be simplified through deterministic sampling of particle drifts with random variations in the initial conditions e.g. \cite{m2013}. However, in practice, forecasting applications are constrained by low resolution models that fail to capture the necessary physical phenomena and as such must be accounted for e.g. through SDEs.

We follow a different approach to account for uncertainty in the drift which does not rely on resolving physics equations. We use a deep neural network (DNN) for modelling the drift (Section \ref{sec:neural-network}), and we propose a probabilistic representation of particle location for representing uncertainty (Section \ref{sec:density-maps}). Through using this representation, our DNN inherently includes the concept of uncertainty in its internal modelling. While we demonstrate our approach using simulated drifts where uncertainty is produced by sampling with respect to the initial position of particles as in~\cite{m2013}, our framework is more general and may be trained with non-simulated drifts or different ways of producing uncertainty.

Our simulations are performed on realistic, high-resolution oceanic surface currents representative of real-world past ocean states in the north-west Mediterranean sea. Thus, we demonstrate our framework by learning a drift model of floating objects at sea. As our framework is completely data-driven, it supports a great deal of flexibility in what it can take in as input. Any information representative of surface currents such as velocity fields or sea surface height (SSH) measures may be used. Even if this information does not capture some of the physical phenomena, a data-driven model may be able to infer the missing phenomena provided they are captured within the training examples. We believe these to be considerable advantages compared to traditional equation-based modelling approaches that lack the flexibility to extract information from different physical quantities and resolutions.

Our contributions may be summarised as: (1) We propose a new approach to modelling Lagrangian drift with uncertainty based on deep learning. Our framework is applicable to observations representative of any source (simulated or not) or measure of uncertainty. (2) This approach is supported by a new statistical representation of particle location for modelling uncertain drift. To the best of our knowledge, no previous framework used a probabilistic location in the modelling of Lagrangian drift. (3) We demonstrate the possibility of modelling Lagrangian drift in scenarios that cannot be solved explicitly through physics equations. (4) We show that performance can be improved by better aligning the learning criteria with the problem setting.

The rest of this article is organised as follows. Section \ref{sec:previous-work} reviews previous works on uncertain drift and trajectory modelling. Section \ref{sec:data} introduces our dataset and Section \ref{sec:method} presents our method. Experimental results are discussed in Section \ref{sec:experiments}. Section \ref{sec:conclusion} concludes the paper.

\section{Previous works}
\label{sec:previous-work}

\subsection{Modelling uncertainty in Lagrangian drift}
\label{sec:previous-work-uncertainty}
Uncertainty in Lagrangian drift is usually modelled using stochastic trajectories through SDEs. Stochasticity may be used to parameterise unresolved physics at subgrid scales, either by formulating the SDE as a Fokker-Planck equation \cite{visser2008} or by fitting an SDE to simulated stochastic trajectories \cite{lacasce2008}. For the application of sea surface currents, examples of unresolved physics are the motions of eddies, waves, or small-scale turbulence. For a review on how to use SDEs to account for oceanic phenomena, see~\cite{van2018}. Stochastic trajectories may be simulated by the means of randomly varying a particle's displacement, velocity, or its acceleration. Contrary to a pure data-driven approach, SDEs may not be able to describe arbitrary sources of uncertainty. Furthermore, they are limited in their reliance on specific physical quantities such as velocity information.

\subsection{Machine learning for Lagrangian drift}
Previous works utilising machine learning to predict Lagrangian drift aim to model drift deterministically rather than probabilistically. \cite{han2021,grossi2020} train their prediction models on individual instances of artificial simulated flows, and as such they do not consider generalisation to different flows (e.g. real spatio-temporal conditions). Instead of using simulations, \cite{nam2020,aksamit2020} learn from past observations of drifters at sea. \cite{nam2020} used a neural network to predict drifter displacement from wind and flow velocity, while \cite{aksamit2020} solve physics equations with a neural network implementing an additional term to learn the unknown physical phenomena. In doing so, \cite{aksamit2020} demonstrated an ability to model drift behaviour that was not described by a baseline physics model. However, such modelling capabilities diminished as the spatio-temporal conditions deviated from the training set, indicating a lack of generalisation to different flows. Due to the limited availability of past observations combined with the passive nature of their drift in real-world environments, the resulting coverage of conditions is typically insufficient to meaningfully represent the true distribution. \cite{zhuang2021} considers the case of learning to model drift of a scalar field as opposed to individual particles. However, their methodology is tightly integrated with the equation for advecting concentration fields, which is fundamentally different to advecting particles with uncertainty.

\section{Data}
\label{sec:data}

We generate a simulated dataset of \textit{probabilistic trajectories} in which an object's position is described by a 2D probability distribution as opposed to a discrete point. In practice, we achieve this by sampling many trajectories in order to estimate the statistics of the underlying uncertainty distribution. Trajectories are sampled by advecting particles on velocity fields of surface currents, which correspond to outputs from a realistic and high resolution numerical ocean model. In this study, we model drift over one day, so we discretise the velocity fields and probabilistic trajectories to one day snapshots. This timestep is motivated by the spatial resolution of the ocean model and the dynamics of the region considered in our study (north-west Mediterranean Sea).

We use modelled surface currents which represent the state of the ocean in our region of study for the years 2018 and 2016. We create one dataset of probabilistic trajectory snapshots per year, ensuring that the entire year is sampled in order to representatively capture any seasonal variances in the currents. The 2018 dataset is used for both training and evaluation\footnote{Due to the limited number of one-day snapshots in a year, the same flow scenarios are used for both training and evaluation for the year 2018. However, trajectories are sampled across different locations (see Section~\ref{sec:trajectories}) which prevents the model from relying on memory.} while 2016 is reserved exclusively for testing. This is to ensure a fair evaluation of our models' ability to generalise to unseen flow scenarios.

\begin{figure*}[t]
    \centering
    \hspace*{\fill}
    \includegraphics[width=0.49\textwidth]{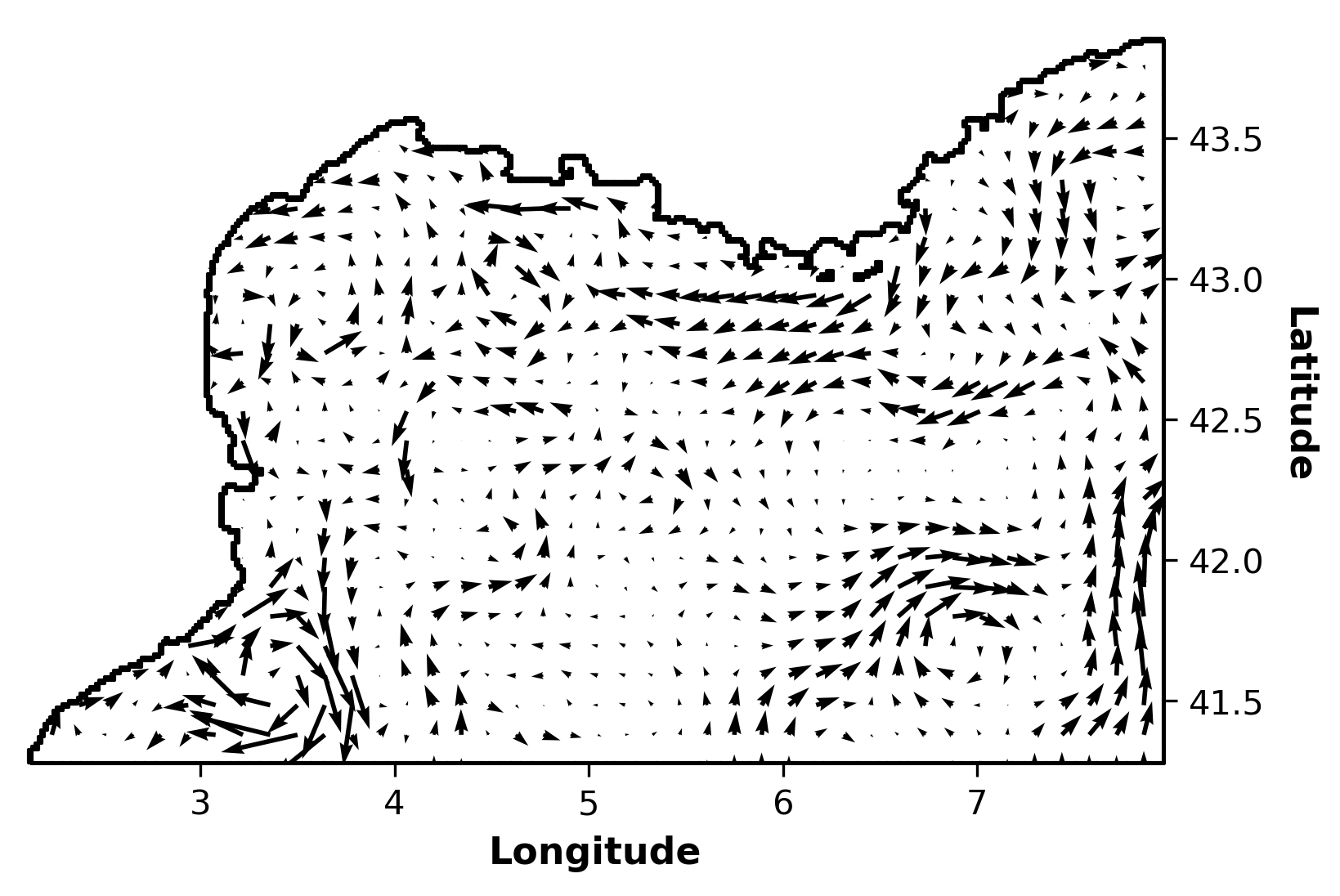}
    \hfill
    \includegraphics[width=0.49\textwidth]{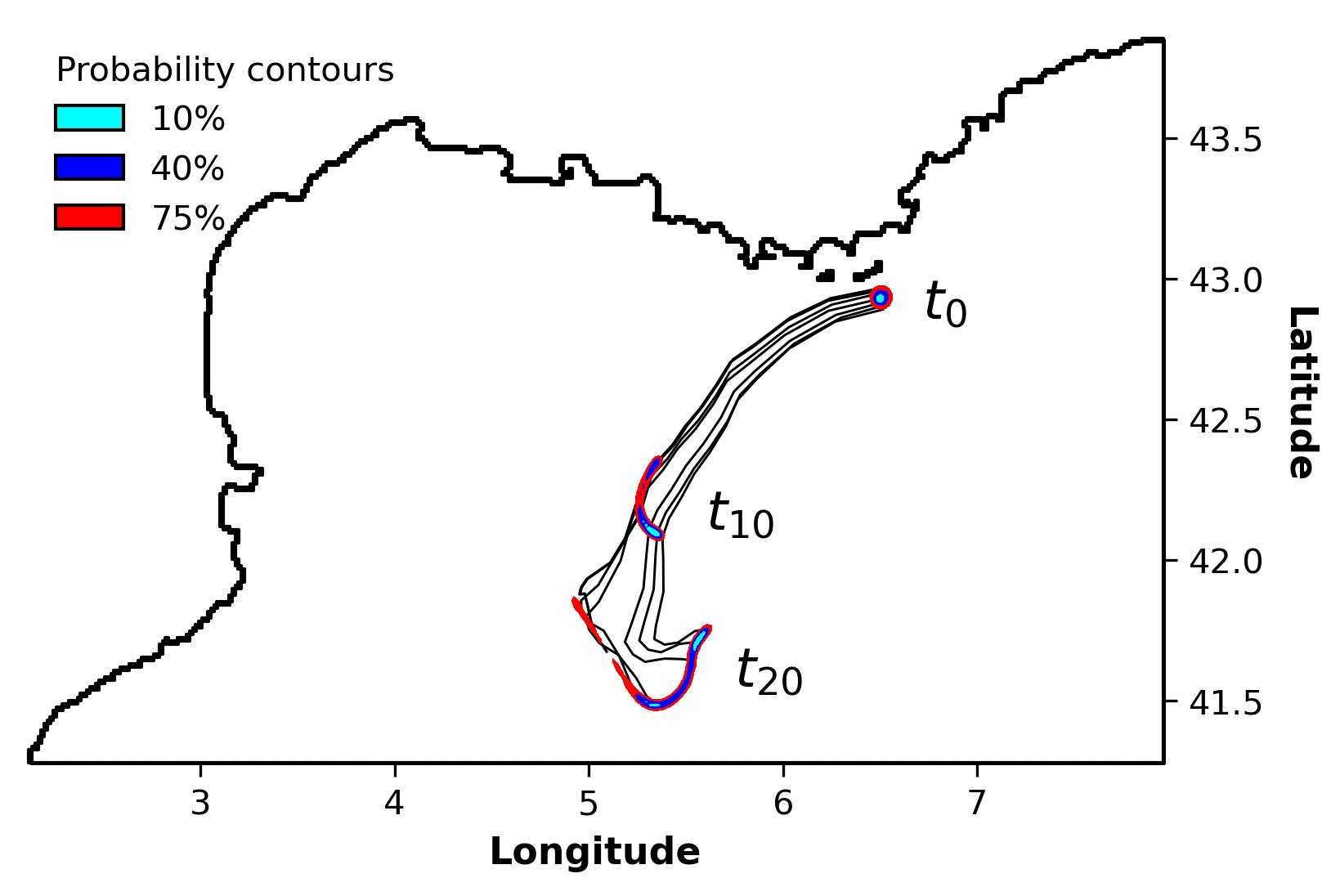}
    \hspace*{\fill}
    \caption{Overview of data. Left) Example velocity field (downgraded resolution for visualisation). Right) Overlay of 3 probability density snapshots (shown as filled probability contours). Example trajectories used to estimate the probability density snapshots are shown in black.}
    \label{fig:data}
\end{figure*}

\subsection{Velocity fields of surface currents}
\label{sec:velocity-fields}

\subsubsection{Ocean model}
We use surface currents from the GLAZUR64 \cite{ourmieres2011} ocean general circulation model (OGCM) which is based on the NEMO \cite{nemo} OGCM. The model has been validated with real observational data (current meters and sea surface temperature/height) \cite{ourmieres2011,guihou2013} in order to provide high-resolution realistic snapshots of past ocean states within the north-west Mediterranean sea (lon 2--8$^{\circ}$ E, lat 41.3--43.9$^{\circ}$ N). In this study, we consider the surface to be two-dimensional by utilising the uppermost layer only. In the ocean modelling community, it is standard to approximate the drift of floating objects by ignoring depth information \cite{mansui2015}. The two-dimensional resolution is \sfrac{1}{$64^\circ$} which equates to each grid cell being representative of $\sim$1.3$\times$1.3 km.

\subsubsection{Data preparation}
GLAZUR64 produces an output every minute for numerical stability purposes, so we average its outputs over a one-day period in order to fulfil our one-day modelling scenario. Its velocity components ($U$ and $V$) are staggered such that $U$ and $V$ are offset by half a grid cell down and to the right, respectively. Prior to giving these components as input to our CNN in Section~\ref{sec:neural-network}, we align the components to the pixel centres using linear interpolation\footnote{Although $U$ and $V$ are gridded on curvilinear coordinates, the limited region that we consider makes it reasonable to neglect projection errors and to associate each grid cell to a Cartesian pixel.}. We also replace the NaN values from land pixels to 0 which provides a natural interpretation of zero flow.

\subsection{Probabilistic trajectories}
\label{sec:trajectories}

Despite the methodological premise of our work being to model drift across a single timestep, we generate long trajectories with the purpose of introducing variance into the level of uncertainty of the snapshots. As a particle traverses over time, the uncertainty in its position will grow as the potential for it to take different paths increases (see Fig.~\ref{fig:data}, right). One-day snapshots are then extracted from the trajectories to define the groundtruth drifts.

\subsubsection{Particle advection}
\label{subsub:advection}
We use the OceanParcels \cite{delandmeter2019} library to advect massless, floating particles on our velocity fields using a 4th order Runge-Kutta integration scheme, where we update the state of the particles every six hours. The positioning of particles is continuous, so OceanParcels performs space-time interpolation of the discretised velocity fields.

We advect particles for up to 15 days and save their positions daily. Particles may not always complete a full 15-day trajectory due to interactions at the boundaries. There are two types of boundaries: the ocean-land boundary and the open boundary (see Fig.~\ref{fig:data}). The open boundary is named as such due to being caused by the cutoff of our data coverage such that the neighbouring ocean values are unknown. We define two conditions for the premature termination of particles: 1) an advection step has caused a particle to escape the ocean-land or open boundary, or 2) a particle has made contact with an ocean cell (pixel) at the open boundary. The second condition exists to prevent particles from getting stuck and accumulating at the open boundary.

\subsubsection{Introducing uncertainty}
\label{subsub:uncertainty}
As discussed in Section~\ref{sec:previous-work-uncertainty}, introducing random behaviour into the modelling of Lagrangian drift serves the purpose of accounting for uncertainty within the data or drift process. Thus, we can estimate the probabilistic nature of the drift conditioned by the underlying uncertainty by sampling many ($N_P$) trajectories that draw from the random distribution. To demonstrate our framework, we sample trajectories whose initial positions draw from a uniform random distribution within a 5 km radius\footnote{If this results in particles to lie outside of the ocean's domain then we discard them, and hence the actual number of particles may be less than $N_P$.}. This choice of randomness is motivated by being simplistic and efficient, as it allows for the advection process to remain completely deterministic. For this type of uncertainty, we empirically observe $N_P$=10,000 as being sufficient for approximating the distribution. In practice, our methodology could be applied to any desired source of randomness such as perturbations in a drifting particle's velocity or position.

\subsubsection{Temporal snapshots of probabilistic trajectories}

In preparation for training our CNN to model probabilistic drift over a given timestep (one day in our experiments), we divide the probabilistic trajectories into temporal snapshots. Each snapshot represents a probability distribution of a particle's position in space. As we approximate this distribution by advecting $N_P$ particles, our snapshots are initially represented as a group of particles, before being converted into probability density maps in Section~\ref{sec:density-maps}. As mentioned previously, particle advection may be prevented at the boundaries, thus the sum of a snapshot's distribution may be less than 1 due to the number of particles in a snapshot being less than $N_P$.

\begin{figure*}
    \centering
    \includegraphics[width=\linewidth]{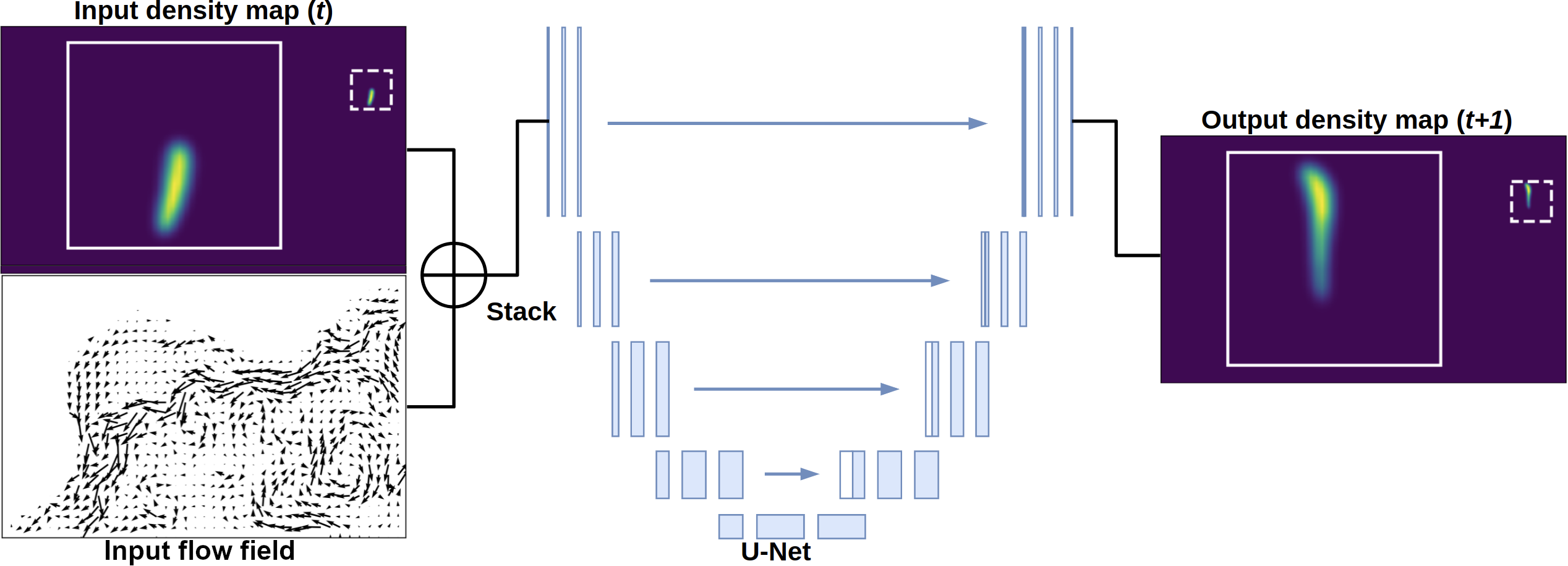}
    \caption{Overview of methodology. The input density map at time $t$ is stacked with the input flow field, which is fed into a U-Net architecture to output the successive density map snapshot at $t$+1. We show zoomed overlays (solid white square) of the foreground region of the density map snapshots (dashed white square). Note that the coastline is not visible in the density maps due to setting the land values to have probability density values of 0.}
    \label{fig:methodology}
\end{figure*}

\subsubsection{Deployment of probabilistic trajectories}

For each dataset, we deploy $N_T$ 15-day probabilistic trajectories whose initial positions are randomly sampled over the ocean's spatio-temporal domain. We choose $N_T$=20,000 to encourage a wide spatio-temporal coverage of our simulated trajectories. The probabilistic trajectories are split into training\footnote{The training split for the 2016 dataset is not used in this study.}, validation, and test sets with a ratio of 70/15/15 prior to discretising them into snapshots to ensure no cross contamination across the sets. To ensure a full 15-day trajectory can be completed, we set the last deployment date to December 16th.

\section{Methodology}
\label{sec:method}

We train a U-Net based CNN to model the probabilistic one-day drift of an object's location. The CNN takes as input a probability density map of object location as well as a representation of the underlying flow field, and outputs the density map of the following day. Our baseline model inputs the true flow field, i.e. the two consecutive velocity fields ($t$~\&~$t$+1) used to simulate the drift in Section~\ref{sec:data}. Thus, our baseline model is an emulator of the simulation environment. Experiments that train models using incomplete flow fields (e.g. missing spatial or temporal information) are given in Section~\ref{sec:exp-imperfect-knowledge}. All models are trained using a pixel-wise regression--based learning criteria, with various criteria considered in~Section~\ref{sec:learning-criteria}.

\subsection{Probability density maps for uncertainty representation}
\label{sec:density-maps}

In Section~\ref{sec:trajectories}, we extracted temporal snapshots from probabilistic trajectories whose distributions were estimated by sampling a large number of particles. Here, we process these snapshots of particles into probability density maps by computing a 2D histogram on their locations with respect to the flow field's grid. This allows us to homogenise the representations between the particle distribution and flow field, which are both given as input into the CNN by stacking them together (see Fig.~\ref{fig:methodology}). To compensate for only having a finite number of particles, we apply a Gaussian filter ($\sigma$=1) to the histogram in order to smooth out its unnaturally sharp spatial gradients. We note that the sum of our density maps may not always sum to one due to the possibility of particles escaping the region, as explained in Section~\ref{subsub:advection}.

\subsection{Neural network}
\label{sec:neural-network}

\subsubsection{Architecture}
While in practice we could use any architecture designed for the context of pixel-wise regression, such as those used for segmentation~\cite{chen2017}~and~\cite{he2017}, we demonstrate our methodology using the popular U-Net architecture~\cite{ronneberger2015}. Fig.~\ref{fig:methodology} shows an outline of the symmetrical encoder-decoder design of U-Net. Each block of the encoder halves its spatial dimensionality and doubles the number of feature channels, while the decoder does the opposite (using bilinear upsampling) in order to gradually recover the output in image space from low-dimensional features. Skip connections are used in the form of concatenating feature maps in order to preserve information at different spatial scales.

We use 64 convolutional channels for the first encoder block, resulting in 1024 feature maps for the final (fifth) encoder block. Unlike the original design, we normalise convolutional outputs using batch normalisation (BN)~\cite{ioffe2015}. Each block of the network therefore consists of two consecutive layers of 3$\times$3 convolution $\rightarrow$~BN~$\rightarrow$~ReLU. Another difference is that we employ residual connections~\cite{resnet} between each block and use the uniform scaling rule as in~\cite{hayou2021} in order to stabilise the gradients. The final layer maps the culminating 64 feature maps into a single channel output using a 1$\times$1 convolution. The bias of this convolution loosely implies the value of the background so we do not include it (i.e. it is frozen at zero).

\subsubsection{Learning criteria}
\label{sec:learning-criteria}
To train CNNs for pixel-wise regression tasks, the most common loss functions used are mean absolute ($\mathcal{L}_1$) and mean squared ($\mathcal{L}_2$)~\cite{carvalho2018}. We try both loss functions in Section~\ref{sec:exp-learning-criteria} to see which is better for our learning task. As we know that probability densities can only exist within the set of ocean pixels $\mathcal{O}$, we ignore any land pixels during the computation of the loss
\begin{equation}
    \label{eq:loss}
    \frac{1}{\vert \mathcal{O} \vert}\sum_{x \in \mathcal{O}} \mathcal{L}(D_x^{t+1}, \hat{D}_x).
\end{equation}

As the advection equation is a partial differential equation (PDE), our learning task can also be seen through the lens of learning to solve a PDE. Given the residual nature of the solutions, reframing the learning task to explicitly learn the residual has been shown to greatly improve the convergence of optimisation~\cite{resnet}. In section~\ref{sec:exp-learning-criteria}, we experiment with learning two types of residual representations. The first uses a long additive skip connection between the input density map $D^t$ and the output of the final 1$\times$1 convolution layer. The loss function remains the same as Eq.~\ref{eq:loss}. The second modifies the loss function to predict the residual map $R$, which is the difference between the target density map $D^{t+1}$ and the input map $D^t$
\begin{align}
    \label{eq:residual-loss}
    \mathcal{L}^R &= \mathcal{L}(R, \hat{R}), \\
    R &= D^{t+1}-D^t.
\end{align}

\subsubsection{Optimisation}
We use the AdamW~\cite{loshchilov2017} optimiser with betas (0.9, 0.999), a weight decay of 1, and an initial learning rate of 1e-4. We decay the learning rate using a cosine annealing schedule~\cite{loshchilov2016} and use the linear warmup strategy as in~\cite{ma2021}, which we apply to both the learning rate and weight decay. We train our models using mixed precision~\cite{micikevicius2017} and a batch size of 24.

\section{Experiments}
\label{sec:experiments}
For all experiments, we train three models with different seeds (0--2) and present the mean and standard deviation of the following metrics:
\begin{itemize}
    \item $\boldsymbol{\mathcal{L}_2}$ \textbf{Mean squared error (MSE)} -- a measure related to the learning criteria which evaluates errors at the local pixel level.
    \item \textbf{IOU}$\boldsymbol{_x}$ -- A geometrical measure which evaluates the intersection over union between the groundtruth and predicted $x\%$ probability contours. A probability contour represents the smallest distribution with a cumulative probability of $x\%$ (as in Fig.~\ref{fig:data}). We define two special cases if the total cumulative probability of the distribution is less than $x\%$. For the groundtruth, the IOU measure is undefined and not included in the statistics. For the prediction, the partial contour encompassing its limited distribution is used.
    \item \textbf{Mass error} -- A global physics-based measure which evaluates the difference between the expected and predicted total displacement of probability density mass, defined as
    \begin{equation}
    \label{eq:mass-error}
        \left\vert \sum_{x \in \mathcal{O}} R_x - \sum_{x \in \mathcal{O}} \hat{R}_x \right\vert .
    \end{equation}
    
\end{itemize}

\begin{table}
\begin{center}
\footnotesize
\setlength{\tabcolsep}{4.5pt}
\begin{tabular}{rcccccc}
\toprule
\vspace{-.3em}

& \thead{\vspace{-1.5em} \\ Zeros \\ MSE \\ (1e-9)}
& \thead{\vspace{-1.5em} \\ Identity \\ MSE \\ (1e-9)}
& \thead{\vspace{-1.5em} \\ Identity \\ IOU$_{50}$ \\ (\%)}
& \thead{\vspace{-1.5em} \\ 50\% contour \\ size \\ (px)}
& \thead{\vspace{-1.5em} \\ Distribution \\ sum \\ (1e-2)}
& \thead{\vspace{-1.5em} \\ Displaced \\ density \\ (1e-2)} \\
\midrule

\textbf{2018} &
$99.15_{71.34}$ &
$136.52_{115.62}$ &
$12.13_{15.05}$ &
$63.03_{62.44}$ &
$92.16_{21.57}$ &
$64.85_{23.74}$
\\

\textbf{2016} &
$100.44_{73.26}$ &
$132.27_{114.96}$ &
$13.36_{15.57}$ &
$60.01_{60.27}$ &
$90.88_{23.36}$ &
$62.58_{24.04}$
\\

\bottomrule
\vspace{0em} % to fix the caption spacing

\end{tabular}
\end{center}
\caption{Groundtruth density map statistics over the 2018 and 2016 validation sets presented as mean$_{\text{std}}$. `Zeros' assumes the prediction is all zeros and `Identity' assumes the prediction is the identity function.}
\label{tab:stats}
\end{table}

\subsection{Learning criteria}
\label{sec:exp-learning-criteria}

\begin{table}
\begin{center}
\footnotesize
\setlength{\tabcolsep}{4.5pt}
\begin{tabular}{ccccccccc}
    \toprule
    & \multicolumn{2}{c}{$\mathcal{L}_2$ (1e-9)} & & \multicolumn{2}{c}{IOU$_{50}$ (\%)} & & \multicolumn{2}{c}{Mass error (1e-2)} \\
    
    \cmidrule{2-3}
    \cmidrule{5-6}
    \cmidrule{8-9}
    
    & 2018 & 2016 & & 2018 & 2016 & & 2018 & 2016 \\
    
    \midrule
    $\boldsymbol{\mathcal{L}_1}$ &
    $2.49_{0.08}$ & $2.52_{0.06}$ & &
    $86.68_{0.22}$ & $86.75_{0.17}$ & &
    $1.63_{0.05}$ & $1.59_{0.03}$ \\ \vspace{.3em}
    
    +$D_t$ &
    $2.47_{0.07}$ & $2.51_{0.08}$ & &
    $86.78_{0.15}$ & $86.85_{0.16}$ & &
    $1.47_{0.03}$ & $1.44_{0.05}$ \\
    
    $\boldsymbol{\mathcal{L}_1^R}$ &
    $2.49_{0.01}$ & $2.54_{0.04}$ & &
    $86.73_{0.06}$ & $86.75_{0.03}$ & &
    $1.46_{0.00}$ & $1.44_{0.03}$ \\ \vspace{.3em}
    
    +$D_t$ &
    $2.63_{0.05}$ & $2.69_{0.06}$ & &
    $86.43_{0.14}$ & $86.43_{0.16}$ & &
    $1.44_{0.01}$ & $1.48_{0.04}$ \\
    
    $\boldsymbol{\mathcal{L}_2}$ &
    $6.67_{0.13}$ & $6.49_{0.11}$ & &
    $77.67_{0.26}$ & $78.24_{0.25}$ & &
    $3.19_{0.06}$ & $3.08_{0.04}$ \\ \vspace{.3em}
    
    +$D_t$ &
    $6.89_{0.05}$ & $6.69_{0.05}$ & &
    $77.59_{0.09}$ & $78.14_{0.10}$ & &
    $2.89_{0.05}$ & $2.85_{0.06}$ \\
    
    $\boldsymbol{\mathcal{L}_2^R}$ &
    $6.91_{0.05}$ & $6.70_{0.06}$ & &
    $77.56_{0.10}$ & $78.13_{0.12}$ & &
    $2.94_{0.03}$ & $2.89_{0.03}$ \\
    
    +$D_t$ &
    $7.37_{0.33}$ & $7.14_{0.32}$ & &
    $76.74_{0.51}$ & $77.33_{0.50}$ & &
    $3.08_{0.08}$ & $3.04_{0.10}$ \\
    
    \bottomrule
    \vspace{0em} % to fix the caption spacing
    
\end{tabular}
\end{center}
\caption{Comparison of different learning criteria evaluated on the validation sets. $\mathcal{L}^R$ indicates regressing the residual map. $+D_t$ indicates adding the input density map to the learned output. All models are trained for 12 epochs.}
\label{tab:results-learning-criteria}
\end{table}

\subsubsection{Loss function}
The results in Table~\ref{tab:results-learning-criteria} show that $\mathcal{L}_1$ loss consistently performs significantly better than $\mathcal{L}_2$ across all metrics. We observe that $\mathcal{L}_2$ struggles to model lower yet important density values, indicating that the value-based exponential term is an unsuitable weighting of importance. This is coherent with the fundamental characteristic of the dataset we generate in Section~\ref{sec:trajectories} in which the level of uncertainty spreads over time. This results in the average density value to decrease as the area of importance grows larger, thereby causing $\mathcal{L}_2$ to favour earlier snapshots whose density values are more concentrated. The lower IOU of the 50\% probability contour demonstrates that this effect is not limited to the tail end of the distribution and indeed extends to critical intervals.

We also observe that unlike $\mathcal{L}_1$, $\mathcal{L}_2$ fails to model the background values as being exactly zero. This has the possibility of biasing the perceived error in some metrics due to the abundance of small values. Mass error, defined in Eq.~\ref{eq:mass-error}, considers the sum over all ocean pixels and thus the totality of the background may be non-negligible. IOU$_x\%$ defines the predicted contour as the entire distribution when the cumulative probability of the groundtruth is at least $x\%$ but the prediction is less than $x\%$, resulting in the presence of many small values to inflate the union and decrease the IOU score. However, we observe that this is rare in practice and does not have a significant impact on the statistics.

\subsubsection{Residual representations}
In Table~\ref{tab:results-learning-criteria}, we find that both forms of residual representations results in a decreased mass error. For $\mathcal{L}_1$ loss, regressing the residual map directly as opposed to adding the input density map to the output also provides the additional benefit of reducing the variance between runs across all metrics.

While the function to be optimised is equivalent for either residual representation, the learning criteria is fundamentally different, and we hypothesise that regressing the residual map provides a more effective criteria due to focusing on the values that change the most across timesteps. This has the effect of the gradient providing information that is explicit for optimising the function that outputs the residual. Otherwise, when regressing the non-residual map, the values that have not changed introduces redundancy and dilutes the training signal. The difference between the two representations is therefore likely to be dependent on the dataset.

Given the statistics in Table~\ref{tab:stats}, the drifts in our dataset appear to have a relatively low degree of redundancy. IOU$_{50}$ is low between $D^t$ and $D^{t+1}$, the amount of displaced density is $\sim$70$\%$ of the distribution sum, and the MSE between $D^t$ and $D^{t+1}$ is greater than the MSE between $D^t$ and zeros.

Table~\ref{tab:results-learning-criteria} also presents results for combining the two residual representations, which in practice has the effect of requiring the negative component of the drift to be scaled by a factor of two. While the performance is only marginally impacted, the measurable difference reinforces the importance of defining a logical learning criteria. Table~\ref{tab:results-different-inputs} shows how this can be even more significant when learning to model more complex tasks.

\subsection{Modelling drift given imperfect knowledge of the flow}
\label{sec:exp-imperfect-knowledge}

\begin{table}[t]
\begin{center}
\footnotesize
\setlength{\tabcolsep}{4.5pt}
\begin{tabular}{rcccccccc}
    \toprule
    & \multicolumn{2}{c}{$\mathcal{L}_2$ (1e-9)} & & \multicolumn{2}{c}{IOU$_{50}$ (\%)} & & \multicolumn{2}{c}{Mass error (1e-2)} \\
    
    \cmidrule{2-3}
    \cmidrule{5-6}
    \cmidrule{8-9}
    
    & 2018 & 2016 & & 2018 & 2016 & & 2018 & 2016 \\
    
    \midrule
    \textbf{Base--12} & 
    $2.49_{0.08}$ & $2.52_{0.06}$ & &
    $86.68_{0.22}$ & $86.75_{0.17}$ & &
    $1.63_{0.05}$ & $1.59_{0.03}$ \\ \vspace{.2em}
    
    $\mathcal{L}^R$ & 
    $2.49_{0.01}$ & $2.54_{0.04}$ & &
    $86.73_{0.06}$ & $86.75_{0.03}$ & &
    $1.46_{0.00}$ & $1.44_{0.03}$ \\
    
    \textbf{Base--18} &
    $2.33_{0.00}$ & $2.38_{0.02}$ & &
    $87.19_{0.06}$ & $87.26_{0.04}$ & &
    $1.51_{0.02}$ & $1.47_{0.01}$ \\ \vspace{.2em}
    
    $\mathcal{L}^R$ & 
    $2.32_{0.01}$ & $2.40_{0.01}$ & &
    $87.26_{0.08}$ & $87.27_{0.06}$ & &
    $1.37_{0.01}$ & $1.36_{0.01}$ \\
    
    \textbf{Base--24} & 
    $2.20_{0.07}$ & $2.30_{0.04}$ & &
    $87.51_{0.23}$ & $87.48_{0.17}$ & &
    $1.45_{0.05}$ & $1.46_{0.02}$ \\ %\vspace{.3em}
    
    $\mathcal{L}^R$ & 
    $2.25_{0.01}$ & $2.37_{0.03}$ & &
    $87.41_{0.07}$ & $87.37_{0.04}$ & &
    $1.37_{0.01}$ & $1.40_{0.02}$ \\
    
    \midrule
    
    \textbf{Fcast--12} &
    $7.86_{0.36}$ & $57.36_{0.17}$ & &
    $75.98_{0.53}$ & $43.20_{0.14}$ & &
    $3.20_{0.11}$ & $3.10_{0.08}$ \\ \vspace{.2em}
    
    $\mathcal{L}^R$ & 
    $8.04_{0.16}$ & $57.44_{0.15}$ & &
    $75.74_{0.24}$ & $43.17_{0.12}$ & &
    $3.12_{0.10}$ & $3.06_{0.06}$ \\
    
    \textbf{Fcast--18} &
    $6.55_{0.12}$ & $57.45_{0.08}$ & &
    $78.07_{0.16}$ & $43.34_{0.04}$ & &
    $2.72_{0.05}$ & $2.71_{0.07}$ \\ \vspace{.2em}
    
    $\mathcal{L}^R$ & 
    $6.62_{0.15}$ & $57.52_{0.24}$ & &
    $78.01_{0.23}$ & $43.36_{0.03}$ & &
    $2.63_{0.04}$ & $2.64_{0.03}$ \\
    
    \textbf{Fcast--24} & 
    $5.84_{0.13}$ & $57.63_{0.39}$ & &
    $79.28_{0.25}$ & $43.38_{0.16}$ & &
    $2.49_{0.05}$ & $2.50_{0.02}$ \\
    
    $\mathcal{L}^R$ & 
    $5.81_{0.07}$ & $57.53_{0.06}$ & &
    $79.40_{0.14}$ & $43.45_{0.05}$ & &
    $2.35_{0.02}$ & $2.42_{0.00}$ \\
    
    \midrule
    
    \textbf{SSH--12} & 
    $11.14_{1.07}$ & $48.58_{0.68}$ & &
    $71.34_{1.43}$ & $46.67_{0.55}$ & &
    $4.56_{0.54}$ & $4.53_{0.51}$ \\ \vspace{.2em}
    
    $\mathcal{L}^R$ & 
    $9.69_{0.55}$ & $51.90_{2.54}$ & &
    $73.13_{0.64}$ & $45.68_{1.02}$ & &
    $3.81_{0.25}$ & $3.86_{0.18}$ \\
    
    \textbf{SSH--18} &
    $8.82_{0.31}$ & $49.73_{1.17}$ & &
    $74.41_{0.46}$ & $46.70_{0.52}$ & &
    $3.66_{0.16}$ & $3.69_{0.13}$ \\ \vspace{.2em}
    
    $\mathcal{L}^R$ & 
    $8.16_{0.26}$ & $51.84_{2.02}$ & &
    $75.39_{0.35}$ & $46.01_{0.88}$ & &
    $3.26_{0.14}$ & $3.39_{0.07}$ \\
    
    \textbf{SSH--24} &
    $8.03_{0.29}$ & $49.36_{1.02}$ & &
    $75.55_{0.44}$ & $46.99_{0.47}$ & &
    $3.36_{0.13}$ & $3.44_{0.11}$ \\
    
    $\mathcal{L}^R$ & 
    $7.29_{0.60}$ & $51.26_{1.90}$ & &
    $76.66_{0.92}$ & $46.41_{0.74}$ & &
    $3.00_{0.28}$ & $3.19_{0.19}$ \\
    
    \bottomrule
    \vspace{0em} % to fix the caption spacing
    
\end{tabular}
\end{center}
\caption{Comparison of modelling scenarios (i.e. input to the network). \textit{Base} refers to inputting the velocity at both $t$ and $t$+1, \textit{Fcast} at $t$ only, and \textit{SSH} at both $t$ and $t$+1 for the variable sea surface height. Results for models trained using both $\mathcal{L}_1$ and $\mathcal{L}_1^R$ are presented, and Model--$X$ indicates the number of epochs the model was trained for.}
\label{tab:results-different-inputs}
\end{table}

\begin{figure}[t]
    \centering
    \includegraphics[width=\linewidth]{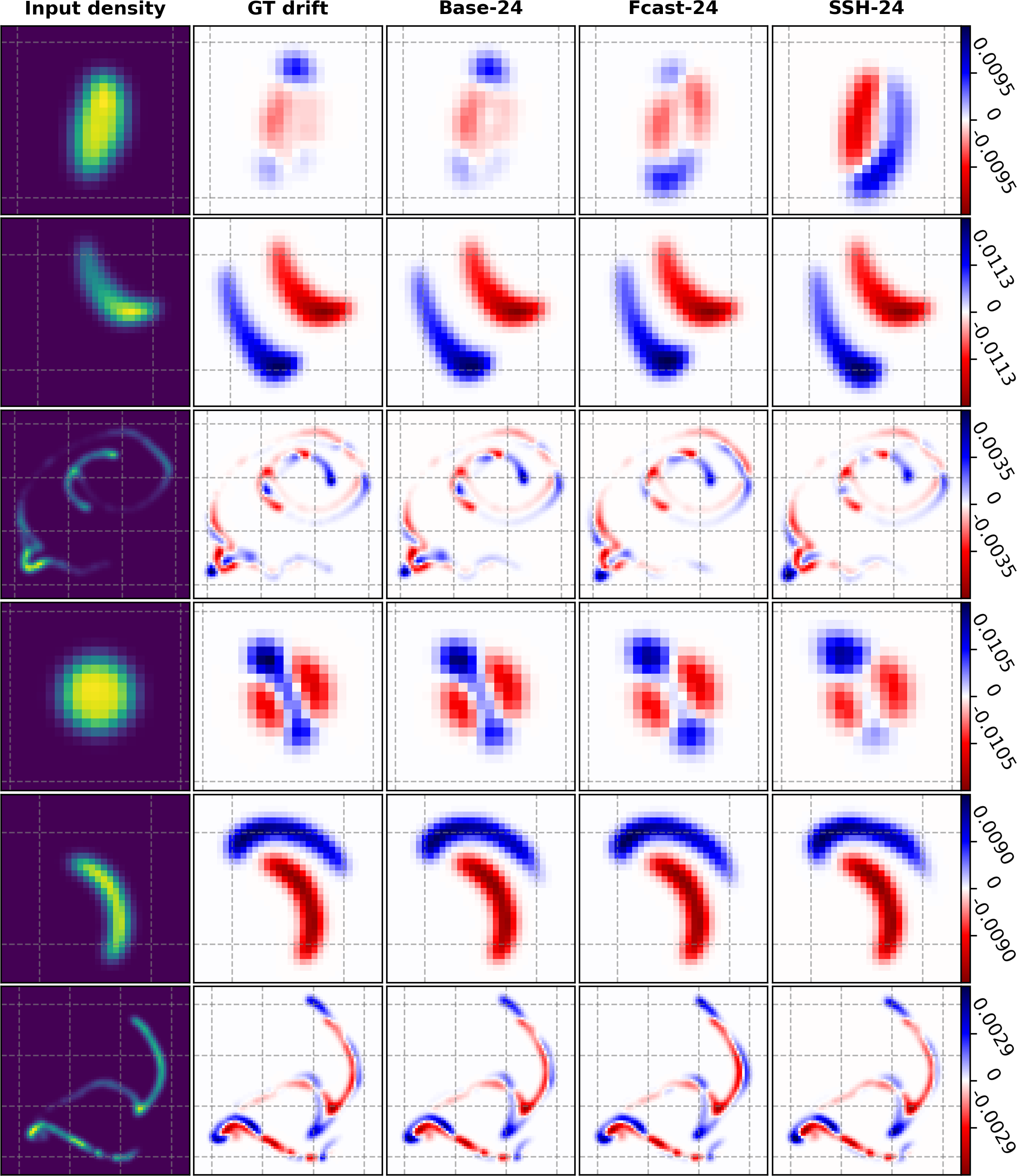}
    \caption{Sample predictions for the three modelling scenarios trained with $\mathcal{L}_1^R$. A range of uncertainty levels are shown, where plots are zoomed on relevant foreground regions (grid scale: 20$^2$ px). Top 3 rows: samples from 2016, bottom 3 rows: samples from 2018. Column 1 shows density values while columns 2-5 show the change of density values between timesteps.}
    \label{fig:results}
\end{figure}

We experiment with two scenarios of learning to model drift given imperfect knowledge. The first (denoted \textit{Fcast}) completely removes information of the velocity at $t$+1, thereby requiring the model to implicitly forecast the velocity's future state. The second replaces information of the velocity field with sea surface height (SSH), a variable which provides significant utility in the field of ocean modelling due to its global availability from satellite observations. It is implicative of sea surface currents at large scales, but does not describe the small scale phenomena that is important for short-term drift modelling. These two scenarios involve learning to recover missing information in the spatial and temporal dimensions, respectively.

We use the exact same architecture and optimisation scheme described in Section~\ref{sec:neural-network} with the only difference being the input to the network. The forecasting scenario inputs a 3-channel matrix ($U_t$, $V_t$, $D_t$) and the SSH scenario inputs a 3-channel matrix (SSH$_t$, SSH$_{t+1}$, $D_t$). Table~\ref{tab:results-different-inputs} presents results for both $\mathcal{L}_1$ and $\mathcal{L}_1^R$, as well as for models trained over different epoch lengths (12, 18, 24).

We observe that the harder modelling scenarios suffer from a generalisation issue when evaluated on a year that was not seen during training (2016). The error for 2016 appears to have a lower bound which is dependent on the modelling scenario, where it neither decreases nor increases as the 2018 validation error decreases (e.g. when training over more epochs). This suggests that the issue is due to data drift rather than overfitting, where the underlying distribution is changing over time. We can also see that the mass error does not change between years, which informs us that the predicted drifts are still physically plausible. With that said, SSH observes an increased lower bound and an increased mass error for 2016 when using $\mathcal{L}_1^R$ as opposed to $\mathcal{L}_1$. This implies that overfitting may still be an issue and that $\mathcal{L}_1$ may help to provide a regularisation effect. Nevertheless, we hypothesise that performance on unseen years could be improved by broadening the distribution represented in the training set by sampling from more years. A direction for future work would be to evaluate the relationship between the number of years represented and the generalisation error.

Empirical observations from Fig.~\ref{fig:results} show that predictions made in the context of the two harder scenarios are generally coherent with respect to large-scale patterns in the groundtruth. However, local details are not always modelled correctly, especially for the unseen year of 2016. For example, columns 1 (2016) and 4 (2018) of Fig.~\ref{fig:results} highlights the difficulties of modelling a local divergence in the flow. Occasionally, the divergence is modelled but with an inaccurate weighting of the displacement, and other times the divergence may not be modelled at all. The latter appears to be more common with SSH. While the issue of generalisation is apparent, it is still worth noting that our models have not reached capacity as Table~\ref{tab:results-different-inputs} shows a steady improvement when training over more epochs. Therefore, it is possible that the aforementioned issues are not fundamental limitations of the modelling scenarios.

\subsection{Validation against artificial flow scenarios}

\begin{figure}[t]
    \centering
    \includegraphics[width=0.8\linewidth]{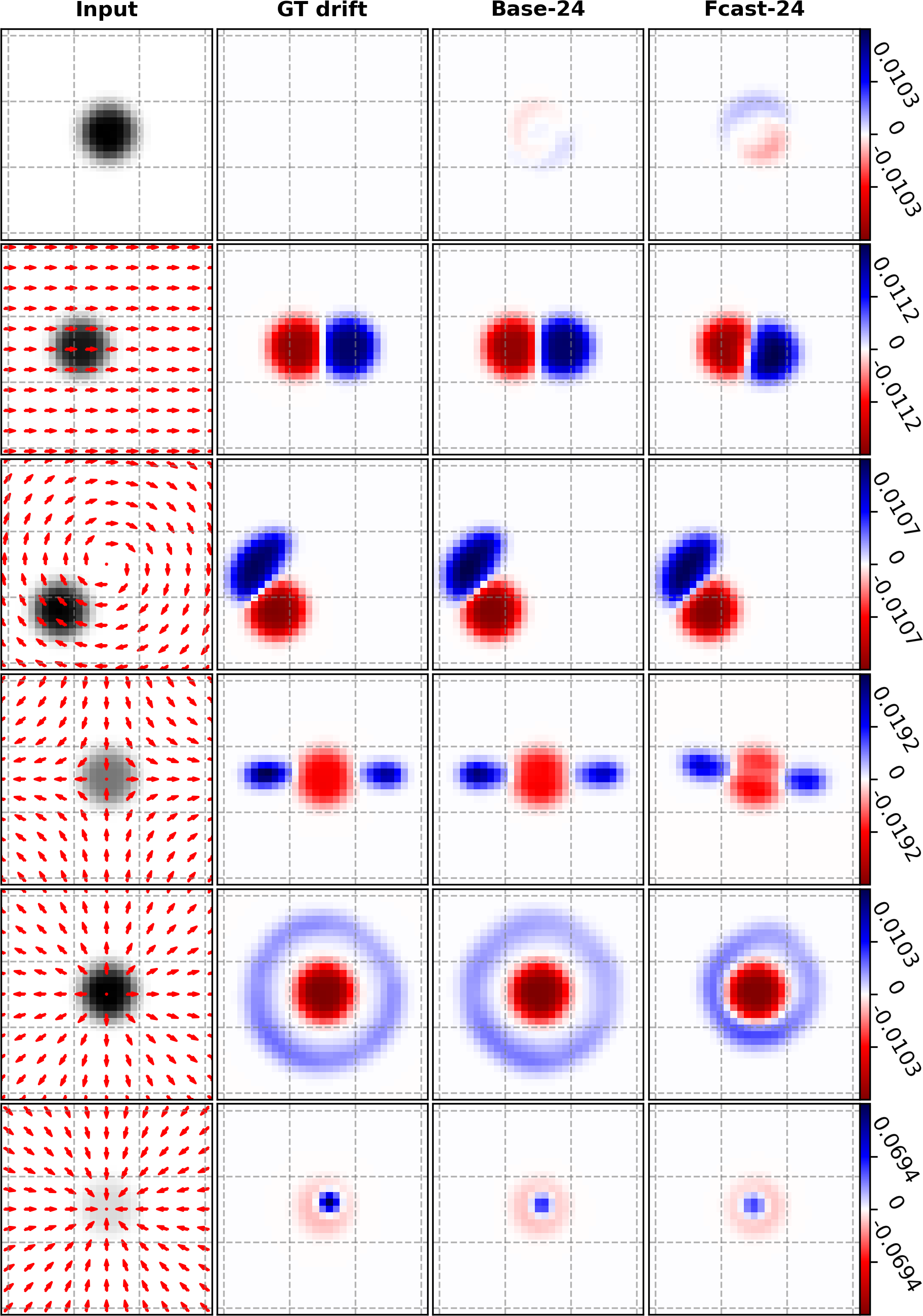}
    \caption{Predictions made over a range of artificial flow scenarios. From top to bottom: zero, constant, vortex, diverging, repelling, and sink. Column 1 shows the superimposed input velocity field and density map. All input density maps are identical but share the same colour scale with the drift maps in order to highlight the relative change in density.}
    \label{fig:stress-test}
\end{figure}

In order to gain insight into the internal biases of our models, we validate them against very simple yet extreme scenarios. These scenarios are static in time, with the velocity magnitude being homogeneous over the field at 0.11 m/s (the mean over the 2018 dataset). Fig.~\ref{fig:stress-test} shows 6 example scenarios for our base and forecasting models. We are unable to validate our SSH model due to the inability to explicitly relate velocity to SSH.

In general, we observe that our base model is able to replicate the expected drift very well despite such extreme scenarios likely being absent in the training set. While the forecasting model is able to model the general idea of the expected flow scenario, we observe that local details are incorrect due to its internal bias of what it expects the next state should be. For example, row 2 of Fig.~\ref{fig:stress-test} presents a very simple scenario where the expected drift is a linear translation along the horizontal axis. We can infer from its prediction that the model expects the velocity field to undergo downwards curvature in the following state. Similar biases can be observed in other examples such as in rows 4 (diverging) and 5 (repelling). While these exact scenarios do not exist in our dataset, the model's inherent pull is likely to be informative of the dataset's average flow evolution in similar contexts. The forecasting model performs well under an artificial vortex scenario likely due to the natural abundance of similar contexts in the dataset.

Row 1 of Fig.~\ref{fig:stress-test} presents a unique scenario in which the input density is expected to remain entirely static due to being presented with a velocity field of zero flow. This scenario only occurs in our dataset for land values, where input density values cannot exist and output density values are excluded from the loss function. We observe that this has the effect of predicting the background as small non-zero values except for the surrounding region of the foreground. The foreground generally appears to remain static, although some degree of drift is observed, with the amount of displacement being 6.6\% and 17.5\% of the input density's maximum value for the \textit{Base} and \textit{Fcast} models, respectively. The respective IOU$_{50}$ scores are relatively high at 96.3\% and 82.8\%.

Row 6 of Fig.~\ref{fig:stress-test} shows an extreme example of the density values being compressed to a very dense region. The maximum value increases by a factor of 6.8, but the \textit{Base} and \textit{Fcast} models are only able to model an increase factor of 2.8 and 2.5, respectively, and in the process loses 35\% and 28\% of the initial mass. As the general idea of compression has been modelled correctly, the loss of mass suggests that the models are sensitive to the absolute density value that can be modelled. 

\section{Conclusion}
\label{sec:conclusion}

We proposed a deep learning framework for modelling Lagrangian drift probabilistically under the influence of uncertainty. Its flexibility allows arbitrary modelling scenarios to be considered, either with respect to the uncertainty distribution being modelled or the underlying flow field being given as input. We demonstrated our framework in the context of modelling floating objects at sea whose initial positions are uncertain. Groundtruth probabilistic drifts were generated by simulating trajectories on surface currents output by an ocean model. We considered three input flow field scenarios for reconstructing the groundtruth drifts: (1) emulator --- surface currents at $t$ \& $t$+1, (2) forecast --- surface currents at $t$ only, and (3) SSH --- sea surface height at $t$ \& $t$+1. We observe that for the harder scenarios of (2) and (3), the models suffer from an inability to generalise well to out of distribution flow scenarios from a different year. Several toy examples of artificial flows helped to give insight into the biases learned from the training distribution for the forecast scenario. We believe that reducing these biases by improving the representativity of the training distribution should be a key focus for future work.

\backmatter

\section*{Acknowledgements}
This work has been supported by Ocean Next, Datlas, and ANRT by means of a PhD CIFRE grant attributed to Joseph Jenkins. It has been co-granted by the FEDER MARITTIMO GIAS (Geolocalisation by AI for maritime Security). Funding was received from Chaire Intelligence Artificielle ADSIL ANR-20-CHIA-0014 and ANR-18-CE40-0014 SMILES. The NEMO calculations were performed using GENCI-IDRIS resources, grant A0110101707.

\bibliography{bib}% common bib file
\bibliographystyle{unsrt}

\end{document}